\documentclass[letterpaper, 10pt, conference]{ieeeconf}
\IEEEoverridecommandlockouts
\usepackage{
    amsbsy, amsfonts, amsmath, amssymb, bm, graphicx,
    multicol, mwe, siunitx, subcaption, xcolor, pdfpages
}
\usepackage[bookmarks=true]{hyperref}
\usepackage{balance}

\title{\LARGE \bf Learning Torque Control for Quadrupedal Locomotion}
\author{Shuxiao Chen$^{*1}$, Bike Zhang$^{*1,2}$, Mark W.\ Mueller$^{1}$, Akshara Rai$^{2}$ and Koushil Sreenath$^{1}$
\thanks{$^{*}$ These authors contributed equally to this work.}
\thanks{$^{1}$ These authors are with the Department of Mechanical Engineering, University of California, Berkeley, CA, 94720, USA, {\tt \{shuxiao.chen, bikezhang, mwm, koushils}\}@berkeley.edu}
\thanks{$^{2}$ These authors are with Meta AI, CA, 94025, USA, {\tt \{bikezhang, akshararai}\}@meta.com}
}

\begin{document}
\maketitle
\thispagestyle{empty}
\pagestyle{empty}

\begin{abstract}
Reinforcement learning (RL) has become a promising approach to developing controllers for quadrupedal robots. Conventionally, an RL design for locomotion follows a position-based paradigm, wherein an RL policy outputs target joint positions at a low frequency that are then tracked by a high-frequency proportional-derivative (PD) controller to produce joint torques. In contrast, for the model-based control of quadrupedal locomotion, there has been a paradigm shift from position-based control to torque-based control. In light of the recent advances in model-based control, we explore an alternative to the position-based RL paradigm, by introducing a torque-based RL framework, where an RL policy directly predicts joint torques at a high frequency, thus circumventing the use of a PD controller. The proposed learning torque control framework is validated with extensive experiments, in which a quadruped is capable of traversing various terrain and resisting external disturbances while following user-specified commands. Furthermore, compared to learning position control, learning torque control demonstrates the potential to achieve a higher reward and is more robust to significant external disturbances. To our knowledge, this is the first sim-to-real attempt for end-to-end learning torque control of quadrupedal locomotion.
\end{abstract}

\section{Introduction}\label{sec:introduction}

Legged locomotion is a longstanding challenge in the field of robotics. While the high degree of freedom, underactuation, and hybrid nonlinear dynamics of legged robots are particularly difficult, they have the advantage of being able to traverse rough terrain, making them highly useful in certain applications. Among the different types of legged robots, quadrupeds have received considerable attention over the last few decades and have achieved significant advancements.

Since the 1980s, model-based control has been a primary method for achieving quadrupedal locomotion \cite{raibert1986legged}. 
This approach uses a dynamics model to develop advanced controllers that enable a quadruped to perform a range of locomotion behaviors with rigorous stability and safety guarantees.
Model-based control for quadrupeds has undergone a significant transformation, shifting from position control \cite{raibert2008bigdog} to torque control \cite{di2018dynamic}. This change has led to more dynamic and robust locomotion behaviors, indicating the superiority of torque control.
On the other hand, learning-based approaches have recently emerged as a promising tool for quadrupedal locomotion, offering great potential for navigating rough terrain \cite{hwangbo2019learning, lee2020learning, miki2022learning}. A notable difference between the learning-based approach and model-based control is that the former typically uses joint positions as the action space, whereas the latter favors torque control. In addition, it is worth noting that learning torque control has been utilized in various robotic applications beyond locomotion, e.g., manipulators \cite{varin2019comparison} and quadrotors \cite{zhang2022zero}, with promising results.
Inspired by the successful transition of model-based control, we explore an alternative approach for learning-based quadrupedal locomotion, where an RL policy directly predicts joint torques, instead of joint positions with a low-level PD controller.

\begin{figure}[t]
    \centering
    \includegraphics[width=\linewidth]{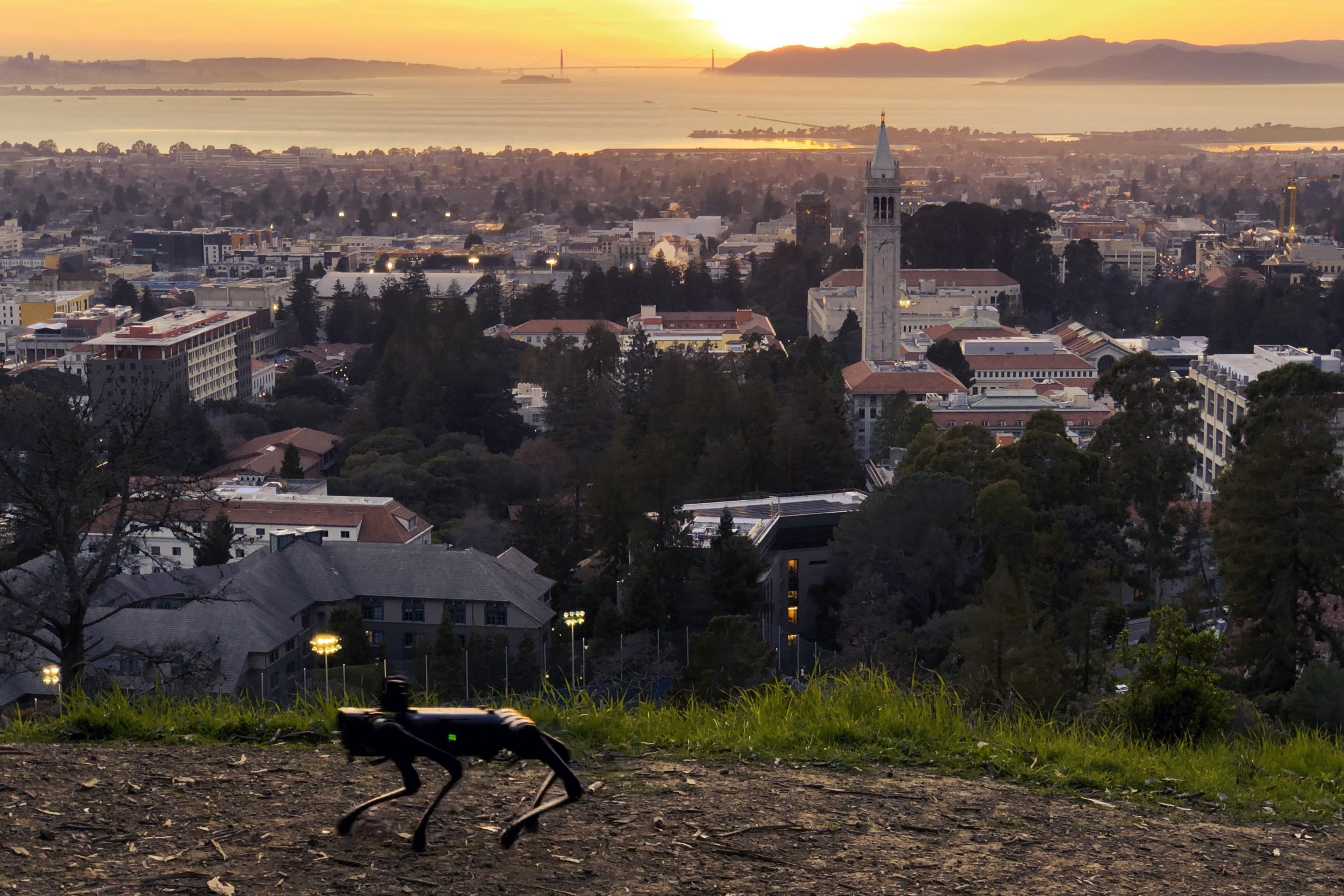}
    \caption{A quadrupedal robot walks on a hill with a torque-based RL policy. The experimental video can be found at \href{https://youtu.be/18v3Z9BD-W8}{https://youtu.be/18v3Z9BD-W8}.}
    \label{fig:cover_picture}
\end{figure}

\subsection{Related Work}\label{sec:introduction:related_work}
Learning quadrupedal locomotion has been widely studied in the past several decades. In the DARPA learning locomotion program, a hierarchical apprenticeship learning method was developed for quadrupedal walking \cite{kolter2008hierarchical}. More recently, deep reinforcement learning has emerged as a viable approach for quadrupedal locomotion. For instance, in \cite{tan2018sim}, a walking policy was trained in simulation and successfully transferred to a real quadrupedal robot. Moreover, imitation from animals was explored to produce learned quadrupedal motions in \cite{peng2020learning}. Several studies have also explored the use of model-based control in learned quadrupedal locomotion, as evidenced by works such as \cite{da2021learning, li2020learning, gangapurwala2020rloc}. In addition to learning only in simulation, a substantial part of learning can also happen in the real world \cite{smith2021legged, wu2022daydreamer}. Furthermore, a series of learning quadrupedal locomotion works have achieved state-of-the-art performance for traversing challenging terrain \cite{hwangbo2019learning, lee2020learning, miki2022learning}. Additionally, other works have demonstrated different learned locomotion behaviors \cite{shi2022reinforcement, bellegarda2020robust, bellegarda2021robust, yang2021fast, choi2023learning} and explored various learning approaches \cite{sun2021online, jiang2021simgan, kumar2021rma}.

For the reinforcement learning-based legged locomotion, there are two main approaches in terms of the action space design: one is joint position \cite{tan2018sim, lee2020learning} and the other is joint torque \cite{lillicrap2015continuous, schulman2015high}. When joint positions are used as actions, the learned policy outputs target joint angles which are then sent to a PD controller. The PD controller is treated as a low-level tracking module, and the PD gains are usually tuned manually to track desired positions well. In addition, a prevalent variation is the residual joint position, where joint positions consist of heuristic or optimization-based joint positions and learned residual joint positions \cite{lee2020learning}. When the action space is joint torque, actions are directly applied to motors to actuate a robot. However, this action choice has only been demonstrated in simulation.

There has been some research to investigate the efficacy of both action choices. A study on different proportional gains for the PD controller was conducted in \cite{xie2021dynamics}, which suggests that high proportional gains can cause instability on the robot while low proportional gains result in a policy behaving like a torque controller that leads to a more robust policy. Nevertheless, low proportional gains have larger tracking errors. In addition, the derivative gain for the PD controller was studied in \cite{smith2022walk}, which validates that a small derivative gain leads to learning instabilities and a large derivative gain prevents the robot from tracking a target velocity. Thus, in order to have adaptive stiffness and damping at the joints, the PD gains can also be learned as part of the action space, as demonstrated in \cite{siekmann2021sim}. However, switching between different PD gains can potentially produce nonsmooth joint torques and cause instabilities of the robot.

In spite of the exploration in the design of PD controller, learning torque control was studied in \cite{peng2017learning} in a simulation setting, where a comparison among different action spaces was conducted, including joint positions, joint torques, joint velocities, and activations for musculotendon units. Note that in \cite[Fig. 4]{peng2017learning}, joint torques are compared with joint positions at the same frequency for a quadruped. However, joint torques need to be executed at a higher frequency in order to achieve comparable performance to joint positions. Moreover, this figure shows that the performance of joint torques becomes better when the query rate increases, which implies the potential of learning torque control. Furthermore, in \cite[Sec. 8]{reda2020learning}, a simulation study of joint positions and joint torques indicates that, while learning joint position control leads to a quicker early learning rate, learning joint position control may converge to lower-reward behaviors and may hinder the ability to adapt to rapid changes.

Building upon the previous studies, in this work, we investigate the possibility of joint torques as the action space for a reinforcement learning policy with application to quadrupedal locomotion. To the best of our knowledge, torque control for real-world quadrupedal locomotion using end-to-end deep reinforcement learning has not been demonstrated yet, and the potential of learning torque control for quadrupedal locomotion is still underexplored.

\subsection{Contributions}\label{sec:introduction:contributions}
The contributions of this work with respect to prior literature are summarized as follows:
\begin{itemize}
    \item We present a sim-to-real learning torque control approach for quadrupedal locomotion, where an RL policy streams joint torques directly, without a PD controller. 
    \item We validate the proposed approach through extensive experiments, in which a quadruped is capable of traversing different terrain and resisting external disturbances.
    \item We illustrate that learning torque control has the potential to achieve higher rewards, compared to the conventional learning position control. 
    \item We demonstrate that a quadruped trained with torque-based RL is capable of resisting larger disturbances when compared with the position-based approach. 
    \item We further discuss the key features of learning torque control and provide implementation suggestions.
\end{itemize}


\section{Learning Torque control for Locomotion}\label{sec:methodology}

\begin{figure*}
    \centering
    \includegraphics[width=\linewidth]{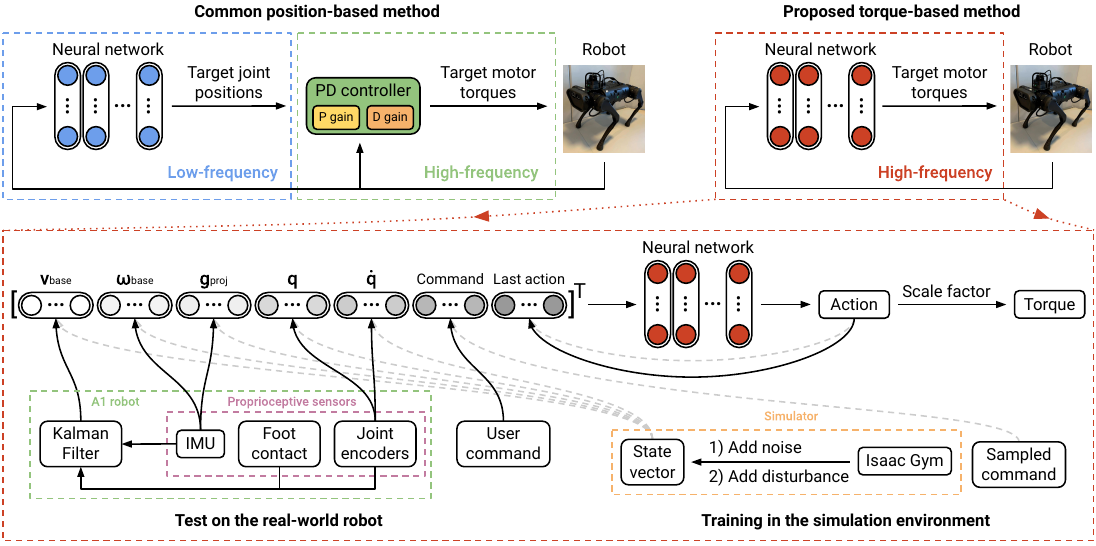}
    \caption{The proposed learning torque control framework. The common position-based method (top left) has two main modules where a neural network (blue) runs at a low frequency and a PD controller (green) executes at a high frequency. The proposed torque-based method (top right) merges the two modules in the position-based method into a high-frequency neural network module (red), as the RL policy is queried at a high frequency to directly output torques. A detailed illustration of the high-frequency neural network module is shown at the bottom, consisting of training in the simulation environment and test on the real quadruped. During the training, we add noise and disturbance to facilitate sim-to-real transfer. During the test, the state information is collected through proprioceptive sensors with appropriate preprocessing. The comprehensive explanation of the framework is presented in Sec.~\ref{sec:methodology}.}
    \label{fig:pipeline_data_flow}
\end{figure*}

In this section, we present our end-to-end learning torque control framework. Fig.~\ref{fig:pipeline_data_flow} illustrates the observations and policy architecture used to predict actions. Compared with the common position-based method, our proposed torque-based policy is queried at a high frequency, and it directly outputs joint torques, avoiding the use of a PD controller.

\subsection{Simulation Environment}\label{sec:methodology:sim_env}

We choose Isaac Gym \cite{makoviychuk2021isaac, rudin2022learning} as our simulator.
We train our policy end-to-end with 4096 parallel environments simulated on an NVIDIA GeForce RTX 3090 GPU. In addition, we use a curriculum of rough terrain in the simulation to improve the robustness of our policy.

\subsection{Problem Formulation}\label{sec:methodology:training:problem_formulation}

We formulate quadrupedal locomotion as a reinforcement learning problem where the objective is to learn a policy $\pi$ that allows an agent to maximize its expected return for a given task.
At each time step $t$, a state $\bm{s}_t$ is observed by the agent from the environment, and an action $\bm{a}_t\sim\pi(\bm{a}_t|\bm{s}_t)$ is sampled from the policy $\pi$. The agent then applies this action and drives its current state to a new state $\bm{s}_{t+1}$ and a scalar reward $r_t = r(\bm{s}_t,\bm{a}_t,\bm{s}_{t+1})$ is received. This process is iterated to generate a trajectory $\tau = \{(\bm{s}_0,\bm{a}_0,r_0),(\bm{s}_1,\bm{a}_1,r_1),\cdots\}$. Therefore, the goal is to learn a policy that maximizes the agent's expected return $J(\pi)$ defined as
\begin{gather*}
    J(\pi)=\mathbb{E}_{\tau\sim p(\tau|\pi)}\left[\sum^{T-1}_{t=0}\gamma^{t}r_t\right]\text{,}
\end{gather*}
where $T$ is the time horizon for each episode, and $\gamma\in[0,1]$ is a discount factor. $p(\tau|\pi)$ represents the likelihood of a trajectory $\tau$ under a policy $\pi$,
\begin{gather*}
    p(\tau|\pi)=p(\bm{s}_0)\prod^{T-1}_{t=0}p(\bm{s}_{t+1}|\bm{s}_t,\bm{a}_t)\,\pi(\bm{a}_t|\bm{s}_t)\;\text{,}
\end{gather*}
where $p(\bm{s}_0)$ is the initial state distribution, and $p(\bm{s}_{t+1}|\bm{s}_t,\bm{a}_t)$ denotes the state transition.

\subsection{Observation Space}\label{sec:methodology:training:observation_space}

We choose a concise representation of the robot that can prove the advantage of learning direct torque control, alleviate overfitting to simulated dynamics, and facilitate sim-to-real transfer \cite{tan2018sim}, excluding any history of proprioceptive sensor measurements or privileged information.
The observation space consists of 7 parts and they together form a $48 \times 1$ vector. The different components consist of robot base linear and angular velocities, orientation, joint positions and velocities, last action, and user commands. Comprehensive definitions of the observation space can be found in Table \ref{table:observation_space}. Base linear and angular velocities and projected gravity are in the robot frame. The projected gravity is a unit vector capturing the robot's orientation with respect to gravity. Desired forward velocity, lateral velocity, and yaw rate are sent to the controller as the user commands to modify the policy behavior online. Joint position, joint velocity, and last action are each 12-dimensional corresponding to the hip, thigh, and calf joints of each of the four legs of the quadruped. The last action is the last neural network output without being multiplied by the action scale.

\begin{table}
    \caption{Detailed definitions of the components forming the observation space. The observation space is divided into 7 observation items whose corresponding symbols are listed side by side. The scale factors stand for the coefficients that are multiplied to the observation items as a preprocessing step. The upper bounds of the zero-mean uniform noise added to the observation are also listed below.}
    \label{table:observation_space}
    \centering
    \renewcommand{\arraystretch}{1.4}
    \resizebox{\linewidth}{!}{%
    \begin{tabular}{|l|c|c|c|}
    \hline
    \multicolumn{1}{|c|}{Observation item} & Symbol & Scale factor & Noise \\ \hline
    Base linear velocity & $\bm{v}_\textrm{base}=[v_{x}\;v_{y}\;v_{z}]^\mathsf{T}\in\mathbb{R}^3$ & 2.0 & 0.2 \\ \hline
    Base angular velocity & $\bm{\omega}_\textrm{base}=[\omega_{x}\;\omega_{y}\;\omega_{z}]^\mathsf{T}\in\mathbb{R}^3$ & 0.25 & 0.2 \\ \hline
    Projected gravity & $\bm{g}_\textrm{proj}=[g_{x}\;g_{y}\;g_{z}]^\mathsf{T}\in\mathbb{R}^3$ & 1.0 & 0.05 \\ \hline
    Joint position & $\bm{q}=[q_1\cdots q_{12}]^\mathsf{T}\in\mathbb{R}^{12}$ & 1.0 & 0.01 \\ \hline
    Joint velocity & $\bm{\dot{q}}=[\dot{q}_1\cdots\dot{q}_{12}]^\mathsf{T}\in\mathbb{R}^{12}$ & 0.05 & 1.5 \\ \hline
    User command & $\bm{u}=[v_{x}\;v_{y}\;\omega_{z}]^\mathsf{T}\in\mathbb{R}^3$ & $[2.0\;2.0\;0.25]^\mathsf{T}$ & 0 \\ \hline
    Last action & $\bm{a}_\textrm{last}=[\tau_1\cdots\tau_{12}]^\mathsf{T}\in\mathbb{R}^{12}$ & 1.0 & 0 \\ \hline
    \end{tabular}}
\end{table}

\subsection{Action Space}\label{sec:methodology:training:action_space}
The action space of our model is a $12 \times 1$ vector consisting of unscaled joint torques. This vector is then multiplied by a constant action scale to produce the actual desired torque values sent to the 12 motors.

The intuition behind the action scaling is as follows: typical position-based RL policies usually add or subtract position values from a reference or nominal position vector, like a standing pose for the robot. There are two reasons for this choice. First, learning a residual on a nominal configuration of joint positions makes it easier for the agent to survive during the early stages of training by giving it a prior standing pose. Secondly, it also helps establish a good starting point for policy exploration because desired gaits often occur around this nominal configuration. 
However, determining nominal torques is difficult for torque control, since these might vary with different environments and initial configurations. Hence, we remove the use of reference or nominal torques and instead use a scaling vector to give final torques to the motor controller.

Due to the fact that the numerical range of torque control (e.g.,~-33 to 33\,\si{Nm} being the safe joint torque range on the A1 robot) is much larger than that of position control (e.g.,~-3.14 to 3.14\,\si{\radian}), we use a large value of $10$ as the action scale. This action scale strikes a good balance between exploration and exploitation in the absence of nominal torques, and it also helps stabilize the robot in the early stage of training. Eventually, all final torques are clamped to the range $[-33,33]$\,\si{Nm} in both simulation and real-world tests to ensure the safety of the robot whilst retaining enough leeway for agile locomotion.

\subsection{Rewards}\label{sec:methodology:training:rewards}
Similar to having no nominal configurations for the action space, no reference trajectories or demonstrations are used in our reward functions either. The learned policy is able to evolve spontaneously from high-level task rewards, and generates gaits that are suitable for high-frequency torque control without any additional constraints to reference motions.

Overall, the total reward function encourages the agent to follow a given set of high-level user commands, whilst satisfying additional behavioral, heuristic, and cosmetic requirements. Detailed definitions of the individual reward terms can be seen in Table \ref{table:reward_terms}. The hip and thigh reward terms drive the positions of the joints close to nominal positions. This creates ``natural looking'' gaits on the robot, enhancing the sim-to-real transfer probability of learned policies.

\begin{table}
    \caption{Detailed definitions of the proposed reward terms. Variables marked with an asterisk stand for target values. $\Delta t$ is defined by the inverse of the control frequency, in this case, 0.005\,\si{\second}. Positive weights mean encouragement to such behaviors and negative ones penalty.
    In the hip (thigh) reward, $H$ ($T$) is the set of 4 hip (thigh) motor indices.}
    \label{table:reward_terms}
    \centering
    \renewcommand{\arraystretch}{1.4}
    \resizebox{\linewidth}{!}{%
    \begin{tabular}{|l|c|c|}
    \hline
    \multicolumn{1}{|c|}{Reward term} & Expression & Weight \\ \hline
    Base linear velocity $x$-$y$ & $\exp{(-\|\bm{v}_{\textrm{base},xy}^*-\bm{v}_{\textrm{base},xy}\|^2/0.25)}$ & $1.0\Delta t$ \\ \hline
    Base linear velocity $z$ & $-v_{\textrm{base},z}^2$ & $-2.0\Delta t$ \\ \hline
    Base angular velocity $x$-$y$ & $-\|\bm{\omega}_{\textrm{base},xy}\|^2$ & $-0.05\Delta t$ \\ \hline
    Base angular velocity $z$ & $\exp{(-(\omega_{\textrm{base},z}^*-\omega_{\textrm{base},z})^2/0.25)}$ & $0.5\Delta t$ \\ \hline
    Orientation & $\|\bm{g}_{\textrm{projected},xy}\|^2$ & $-1.0\Delta t$ \\ \hline
    Joint torque & $\|\bm{\tau}\|^2$ & $-0.0002 \Delta t$ \\ \hline
    Joint acceleration & $\|\bm{\dot{q}_\textrm{last}-\dot{q}}\|^2$ & $-0.00000025\Delta t$ \\ \hline
    Feet air time & $\sum^{4}_{i=1} (t_{\textrm{swing},i}-0.5)$ & $2.0\Delta t$ \\ \hline
    Collision & $\sum^{4}_{i=1} c_{\textrm{knee},i}$ for $c_{\textrm{knee},i}\in\{0,1\}$ & $-1.0\Delta t$ \\ \hline
    Action rate & $\|\bm{a}_\textrm{last}-\bm{a}\|^2$ & $-0.01\Delta t$ \\ \hline
    Hip & $\sum_{h\in H} |q_h^*-q_h|$ & $-1.0\Delta t$ \\ \hline
    Thigh & $\sum_{h\in H} |q_t^*-q_t|$ & $-1.0\Delta t$ \\ \hline
    \end{tabular}}
\end{table}

\subsection{Policy Architecture and Training}\label{sec:methodology:training:model}
The policy is modeled with a multilayer perceptron (MLP) which has 3 layers of 512, 256, and 128 exponential linear units (ELUs). The standard deviation of the action distribution is set to a fixed value. The value function is modeled by a separate network with the same configuration as the policy network. The policy is optimized with proximal policy optimization (PPO) \cite{schulman2017proximal}. Note that we do not choose a complicated architecture, such as a teacher-student structure in \cite{lee2020learning}, since we would like to use a simple architecture to validate our learning torque control framework.

\subsection{Domain Adaptation}\label{sec:methodology:training:adaptation}

\begin{figure*}
    \centering 
    \begin{subfigure}[t]{0.245\linewidth}
        \centering
        \includegraphics[width = 1.0\linewidth]{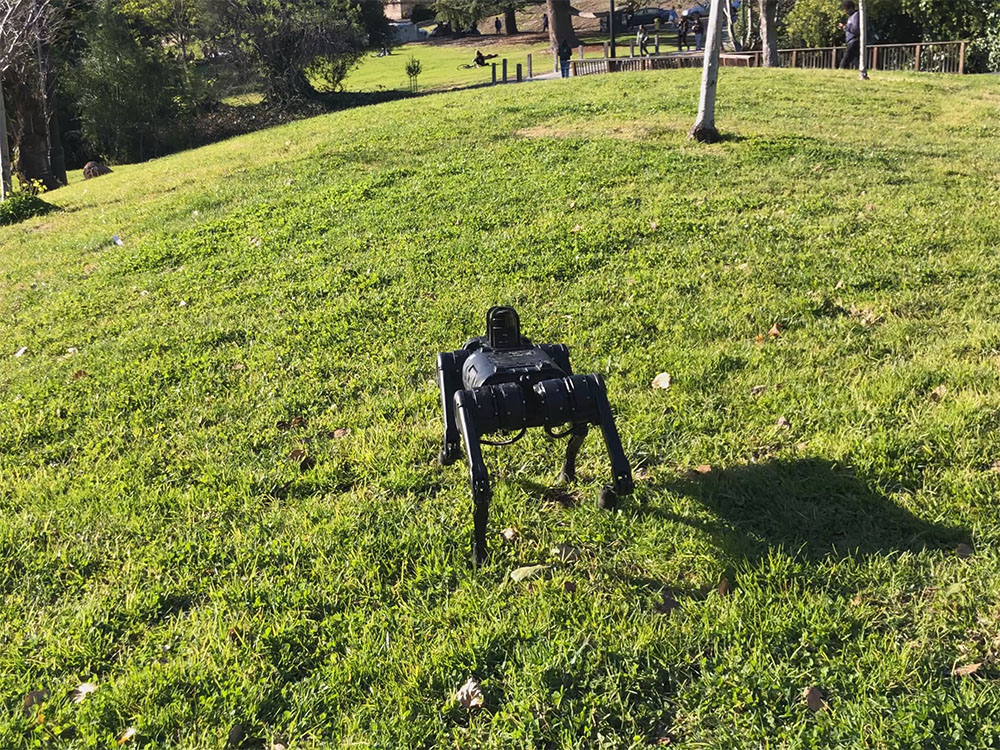}
        \caption{}
        \label{subfig:task_1}
    \end{subfigure}
    \begin{subfigure}[t]{0.245\linewidth}
        \centering
        \includegraphics[width = 1.0\linewidth]{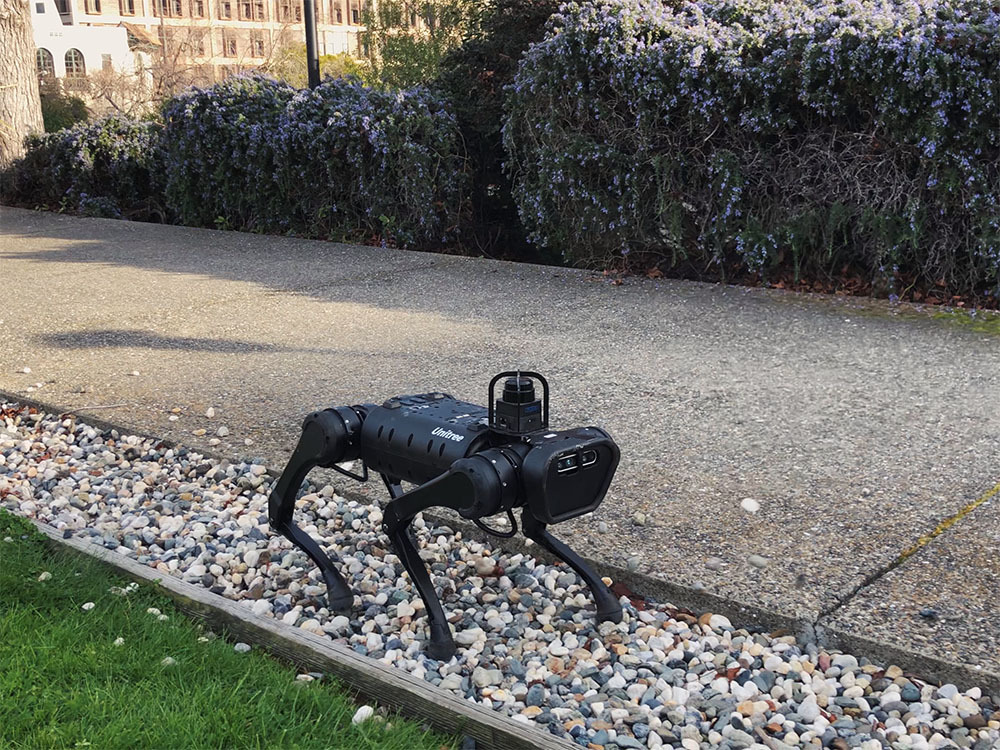}
        \caption{}
        \label{subfig:task_2}
    \end{subfigure}
    \begin{subfigure}[t]{0.245\linewidth}
        \centering
        \includegraphics[width = 1.0\linewidth]{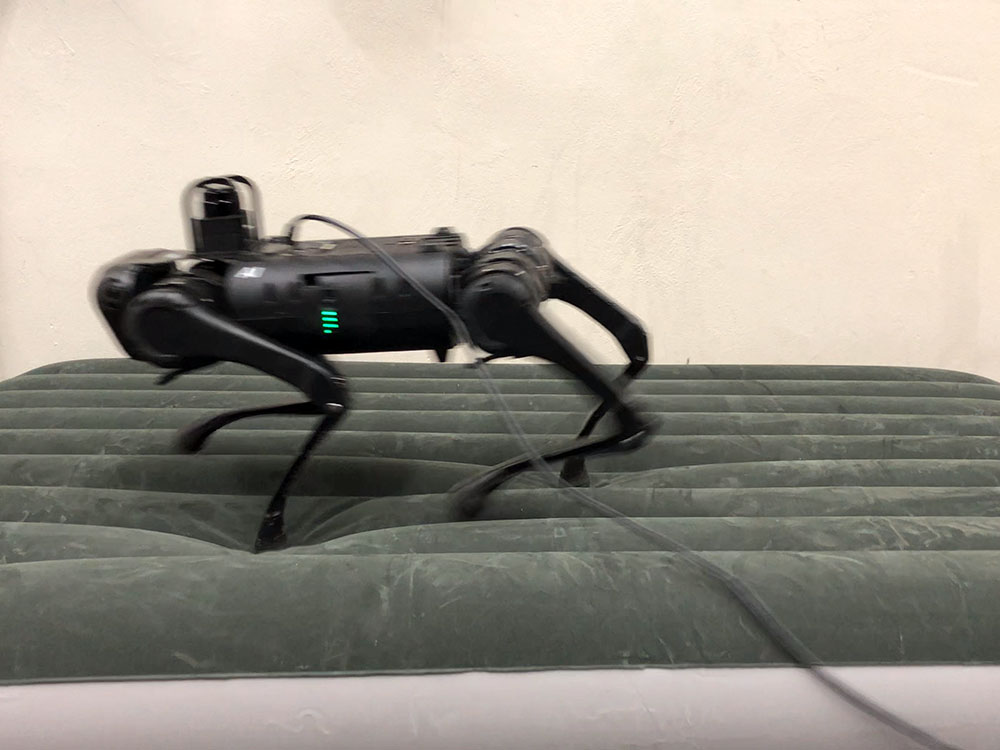}
        \caption{}
        \label{subfig:task_3}
    \end{subfigure}
    \begin{subfigure}[t]{0.245\linewidth}
        \centering
        \includegraphics[width = 1.0\linewidth]{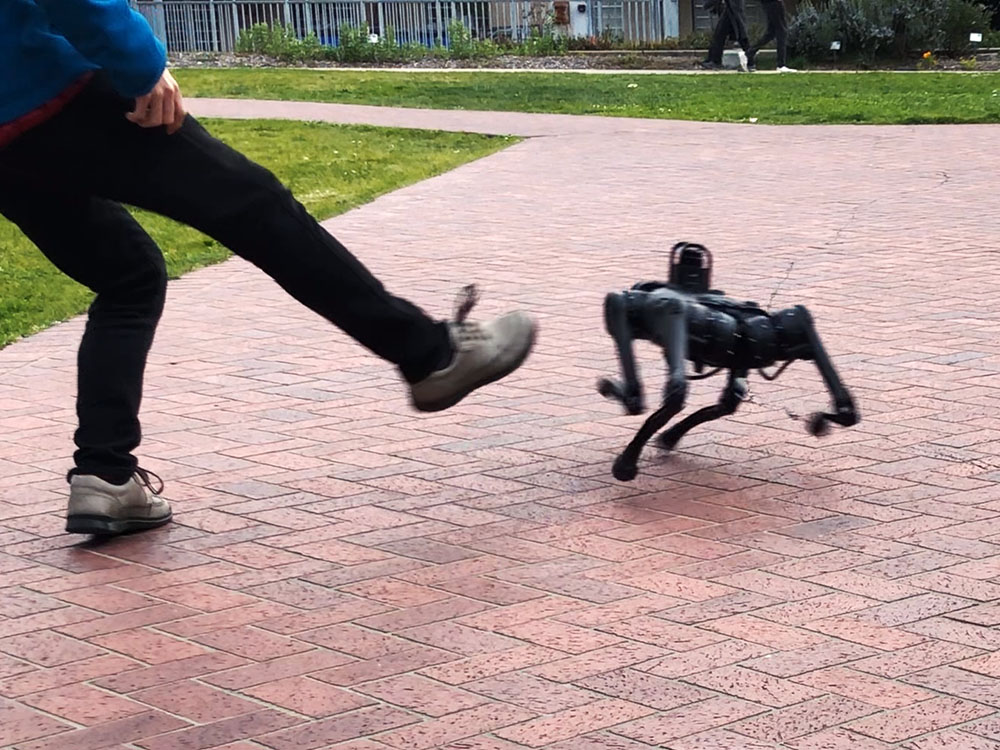}
        \caption{}
        \label{subfig:task_4}
    \end{subfigure}
    \caption{The proposed end-to-end learning torque control framework allows the A1 robot to traverse various terrains, such as (a) inclined soft lawn, (b) pebble path, (c) compliant terrain, and resist external disturbances such as (d) recovering from a kick. All the tests use the same RL policy queried at 200\,\si{\hertz}.}
    \label{fig:various_tasks}
\end{figure*}

Randomization of ground frictions and external disturbances is included to facilitate the building of robustness. A friction coefficient is sampled uniformly in the range $[0.5,1.25]$ and assigned to each simulated robot. External pushes are exerted on each robot every 15\,\si{\second} by sampling and setting the instantaneous $x$-$y$ velocity of the robot base from $[-1.0,1.0]$\,\si{\meter/\second}. Note that uncertainties of the physical parameters, e.g., dynamics randomization \cite{antonova2017reinforcement, peng2018sim}, can be viewed as external forces applied to the robot \cite{pinto2017robust, tan2018sim}. The linear velocity commands are also randomized in range $[-1.0,1.0]$\,\si{\meter/\second} in both $x$- and $y$-directions, and the yaw rate is randomized in range $[-3.14,3.14]$\,\si{\radian/\second}, in order to produce a goal-conditioned RL policy.

\section{Experimental Setup and Results}\label{sec:exp_results}

Having presented the end-to-end learning torque control framework, we now evaluate it with various experiments. 
We first introduce the experimental apparatus and a baseline policy. Then, real-world experiments and comparisons are demonstrated. Finally, we provide an in-depth discussion on the learning torque control framework. 

\subsection{Apparatus}\label{sec:exp_results:apparatus}

We use a Unitree A1 quadrupedal robot to evaluate our end-to-end learning torque control framework. The A1 robot has 18 degrees of freedom (DoFs) with 12 actuated joints and 6 generalized coordinates for the floating base. There is an NVIDIA Jetson TX2 inside the robot body to offer extra computing power. The TX2 has a 256-core GPU for neural network inference implemented with Mobile Neural Network (MNN) \cite{alibaba2020mnn}, an open-source inference engine that is lightweight and supports various hardware backends. In addition, our policy runs completely onboard.

\subsection{Baseline}
We develop a position-based RL policy for the same quadruped as a baseline. This policy uses joint positions as the action space and runs at 50\,\si{\hertz}, followed by a low-level PD controller tracking at 200\,\si{\hertz}, which is the same frequency as the torque-based RL policy. For the PD controller, the proportional gain is 20, and the derivative gain is 0.5. This position-based RL policy is trained with the same level of noise and disturbance as well as the terrain curriculum as the torque-based RL policy. In order to make a fair comparison, we make the reward function of the position-based RL close to the torque-based RL one. However, the same reward function does not work directly for position-based RL policy. We modify the reward function in a minimal way which excludes the orientation, hip, and thigh reward terms, and changes the weight of the feet air time reward from 2.0 to 1.0. The action scale for this position-based policy is 0.25 with a nominal position. This position-based policy also allows a quadruped robot to traverse different terrains.

\subsection{Results}\label{sec:exp_results:results}

We conduct experiments both in simulation and on hardware to extensively study our proposed learning torque control framework. In the simulation, we compare the difference between learning torque control and learning position control in terms of their learning curves. On hardware, we validate the learned torque control policy on various terrains with user-specified commands. The snapshots of the learned locomotion behaviors are shown in Fig.~\ref{fig:cover_picture} and Fig.~\ref{fig:various_tasks}.

In Fig.~\ref{subfig:task_1}, a quadruped robot walks on an inclined soft lawn on campus, which shows that the learned torque policy allows the quadruped to locomote over non-flat terrain while following commands. In Fig.~\ref{subfig:task_2}, the quadruped robot walks on a pebble path, where the pebbles are not embedded in the ground and can be slippery. While the robot walks forward, the pebbles collide with the feet, which causes additional disturbances to the robot. In this experiment, the robot gets stuck in pebbles for a moment. The operator increases the forward velocity command, and then the robot is able to get out of the pebbles and keeps walking forward. Fig.~\ref{subfig:task_3} demonstrates that our torque policy enables the quadruped robot to walk over a compliant terrain, i.e., an air mattress. A similar test was conducted in \cite{choi2023learning} using a position-based RL policy with an adaptive control architecture. In addition to walking on various terrains, we also test the robot with external disturbances as shown in Fig.~\ref{subfig:task_4}. We kick the robot from different angles when it is standing or walking, and it is capable of resisting a certain amount of external forces. All of these locomotion behaviors are performed with the same torque-based RL locomotion policy queried at 200\,\si{\hertz}.

\begin{figure}
    \centering
    \includegraphics[width=1.0\linewidth]{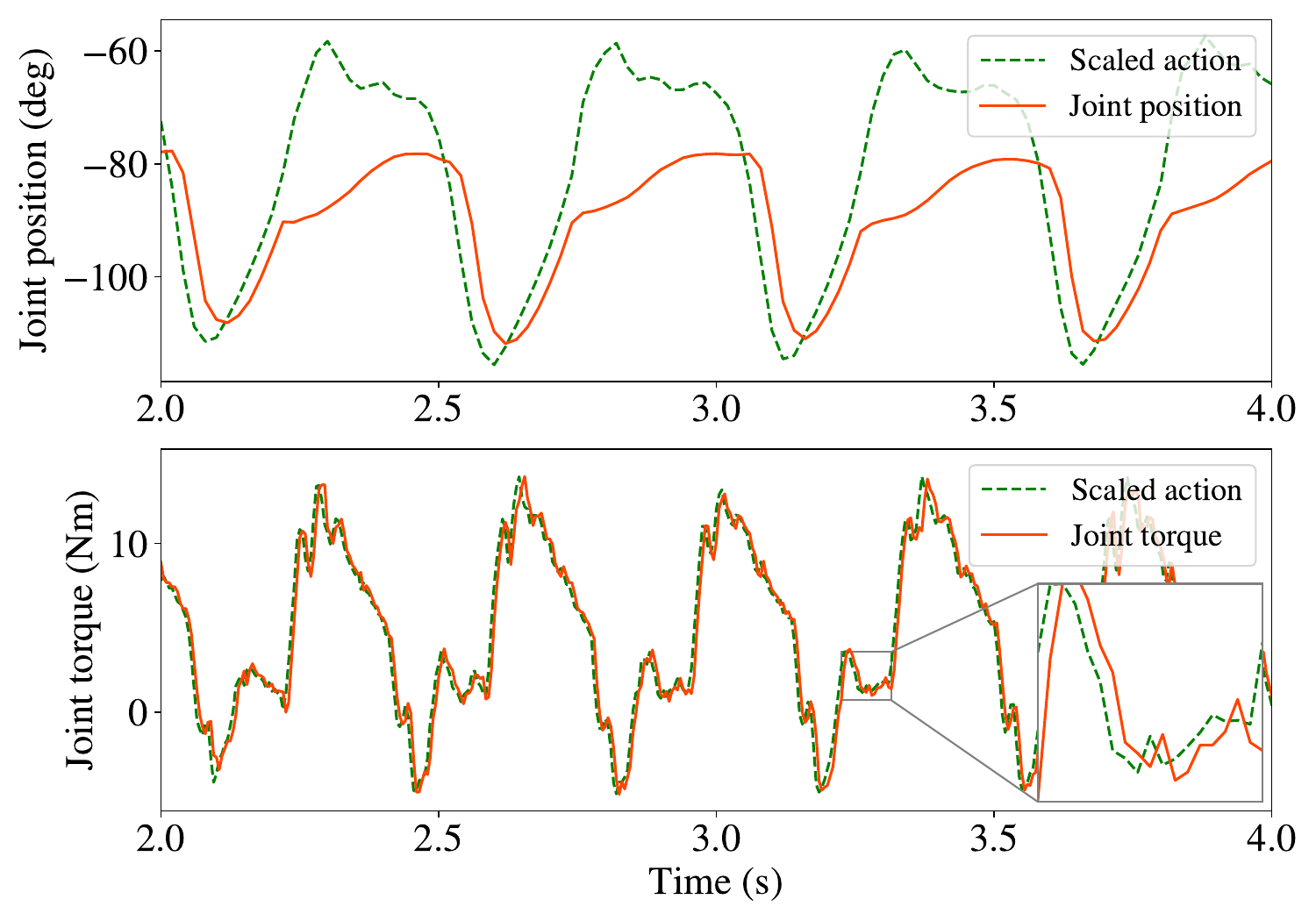}
    \caption{\textbf{Action tracking comparison} between learning position control (top) and learning torque control (bottom) on a real quadruped robot for the rear left calf joint. The green lines depict the target values queried from RL policies with post-processing, e.g., scales. The orange lines illustrate the actual values measured from the real quadruped robot.}
    \label{fig:action_tracking}
\end{figure}

In order to study the difference between learning position control and learning torque control, we carry out real-world experiments to compare their action tracking performances. We perform real-world experiments on the same A1 robot given the same forward velocity command at 0.5 m/s. The action tracking result for the rear left calf joint is shown in Fig.~\ref{fig:action_tracking}, where the top plot illustrates the result of learning position control and the bottom plot demonstrates the result of learning torque control. The green lines denote the target values produced from RL policies with post-processing, and the orange lines indicate the actual values measured from the hardware. Learning position control has a larger tracking error, whereas learning torque control has a much better tracking performance. It is worth noting that this comparison does not indicate that learning position control has worse performance, since the position-based RL can adjust the target position so that the actual position achieves the goal during training. Here, we only aim to demonstrate a major difference between the two methods.

We were concerned that joint torques need to be changed rapidly between the stance phase and the swing phase for a quadruped robot, compared to joint positions. Thus, it may pose challenges to RL to learn to control a quadruped robot using joint torques as the action space. Fig.~\ref{fig:action_tracking} indicates that RL is powerful enough to handle this rapid change in the action space. In addition, we were concerned that the action of RL might not be smooth enough to be fed to motors. For learning position control, there is a low-level PD controller to track the action of RL, and it can locally smooth out jerky values. Fig.~\ref{fig:action_tracking} demonstrates that torque-based RL produces smooth values for motors. Although we occasionally notice some jerky values, the motors on the hardware are generally tolerant. There is no need to add another layer, e.g., a PD controller or a low pass filter, to smooth out the RL actions.

\begin{figure}
    \centering
    \includegraphics[width=1.0\linewidth]{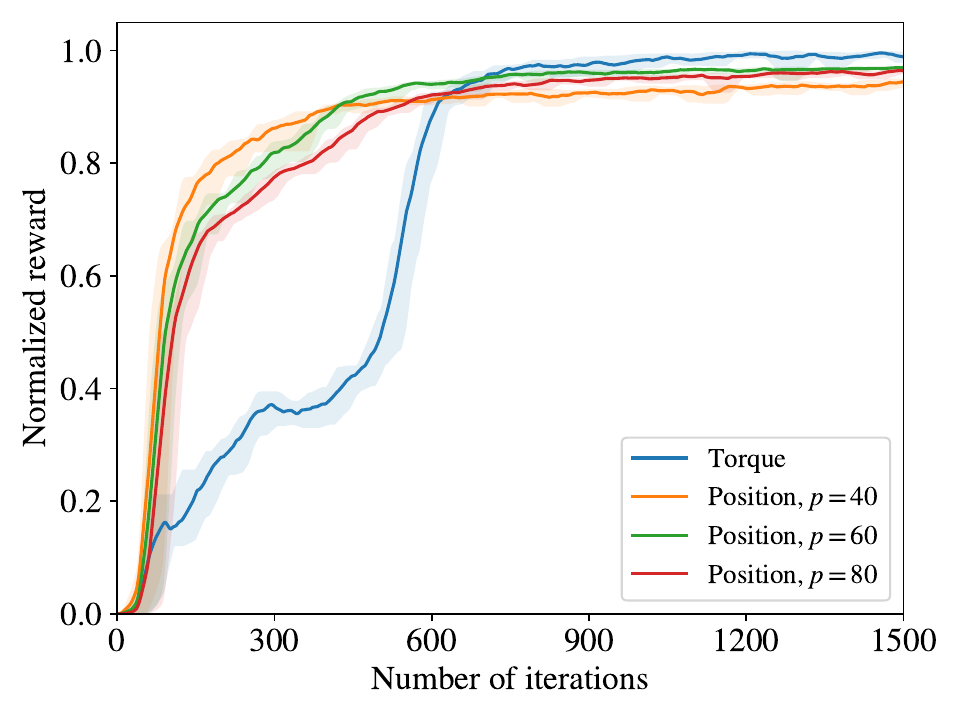}
    \caption{\textbf{Learning curve comparison} between learning position control and learning torque control. We choose three different proportional gains, i.e., 40, 60, and 80, for the position-based RL, which are denoted as orange, green, and red, respectively, and the torque-based RL is depicted as blue.}
    \label{fig:learning_curve}
\end{figure}

Furthermore, we compare the normalized learning curves for learning position control and learning torque control, as shown in Fig.~\ref{fig:learning_curve}. We simplify the reward function design for this comparison, which only consists of the base velocity tracking reward. We choose three different proportional gains, i.e., 40, 60, and 80, which are illustrated as orange, green, and red lines, and the derivative gain is 0.5. The torque-based RL is depicted as the blue line. 
The learning curves for the position-based RL change with respect to different proportional gains, which indicates that proportional gains have an impact on the learning process. 
In contrast, while torque-based RL shows a slow learning rate at the early stage, it converges to a higher return, which shows that torque-based RL has the potential to achieve a higher expected return, compared to position-based RL. This result coincides with the findings presented in \cite{reda2020learning}, which conducts a similar comparison but using a much simpler system.

\begin{figure*}
    \centering 
    \begin{subfigure}[t]{\linewidth}
        \centering
        \includegraphics[trim={1.2cm 21.5cm 7cm 1.6cm},clip,width=\linewidth]{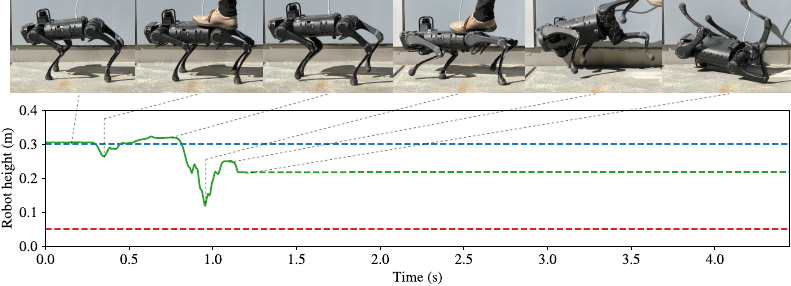}
        \caption{When using the position-based RL policy, the robot is able to sustain itself during the first gentle kick, but after the second larger kick, the robot falls over. The green dashed line stands for the last valid base height measurement before the flip.}
        \label{subfig:large_kick_position}
    \end{subfigure}\vspace{2.5ex}
    \begin{subfigure}[t]{\linewidth}
        \centering
        \includegraphics[trim={1.2cm 16cm 7cm 6.8cm},clip,width=\linewidth]{figures/kick_down_v2.pdf}
        \caption{When using the torque-based RL policy, the robot is capable of resisting three consecutive kicks successfully, and the third kick is the strongest one during which the robot base even contacts the ground.}
        \label{subfig:large_kick_position}
    \end{subfigure}
    \caption{\textbf{Robustness comparison} between learning position control and learning torque control, where the robot base height is illustrated under successive downward kicks while it is standing. The green lines show the measured base heights, and the blue lines depict the nominal base heights. The red lines illustrate the minimal base heights where the base is on the ground. Note that these vertical kicks are not present during training.}
    \label{fig:large_kick}
\end{figure*}

In addition, we compare the robustness between the position-based RL policy and the torque-based RL policy. We perform a real-world experiment where the robot stands while getting consecutive kicks, and the result is presented in Fig.~\ref{fig:large_kick}. The green, blue, and red lines show the measured, nominal, and minimal base heights, respectively. The robot base heights are calculated by averaging the $z$-direction foot positions in the robot frame obtained by forward kinematics (FK). Since the robot body has non-negligible thickness, the minimally possible base height measured from FK is \SI{0.05}{\meter} when the robot body touches the ground. 
In the top plot, a position-based RL policy allows the robot to resist the first kick, but the robot falls over after the second kick. In this case, the base height is no longer valid after \SI{1.2}{\second} due to the feet's loss of contact to the ground.
In the bottom plot, with a torque-based RL policy, the robot is able to sustain itself during three successive kicks. Note that the third kick is the hardest one where the quadruped base even contacts the floor.
This kicking experiment demonstrates that the torque-based RL policy shows more robustness than the position-based RL policy.
We hypothesize that such robustness comes from torque control.
Note that this robust behavior is different from the learned recovery behavior in \cite{lee2019robust}, as we do not explicitly train the robot to resist vertical kicks.
The robustness of the position-based RL policy can be further improved if we explicitly train with downward kicks.

\subsection{Discussion}\label{sec:exp_results:discussion}

\subsubsection{Control Frequency}
Our torque-based RL policy is queried at 200\,\si{\hertz}, whereas the query frequency of most position-based RL policies ranges from 20 to 100\,\si{\hertz} \cite{lee2020learning}, followed by a low-level PD controller. During our preliminary test of the policy querying frequency, the GPU on the TX2 can stably infer the 3-layer MLP and maintain the inference at 20\,\si{\kilo\hertz} without executing its outputs on the robot hardware, which indicates the potential for a higher-frequency policy if needed. Moreover, compared to the position-based RL policy, the minimum query frequency of the joint torque-based policy is supposed to be higher.

\subsubsection{Sim-to-Real Transfer}
One of our concerns in learning torque control was the difficulty of transferring a torque-based policy from simulation to hardware. Surprisingly, it turns out that there is no additional difficulty of sim-to-real transfer for learning torque control, compared to the conventional learning position control. The common domain randomization strategy leads us to successful sim-to-real transfer for both position-based and torque-based policies.

\subsubsection{Circumvention of PD Controllers}
Learning torque control allows us to circumvent the use of a low-level PD controller, as the torque-based RL policy directly outputs joint torques. For the position-based locomotion policies, PD gains are usually tuned manually and tightly related to the learning locomotion performance, as studied in \cite{xie2021dynamics, smith2022walk}. Thus, the torque-based RL method eliminates the number of parameters to tune. Especially, these PD gains need to be tuned on hardware involving extra engineering work.

\subsubsection{Potential of Learning Torque Control}
This is the first attempt, to our knowledge, of learning torque control for quadrupedal locomotion with real-world experiments. From a perspective of biomechanics, torque control is more inclined to the way that animals or humans locomote \cite{adolph2012you, han2021impact}. Hence, we hypothesize that learning torque control can potentially lead to more dynamic and flexible behaviors in the future.

\section{Conclusion}\label{sec:concl_future}
In this paper, we studied an alternative approach to learning to control a quadruped robot, which uses joint torques as the action space in reinforcement learning, instead of joint positions as conventional learning quadrupedal locomotion work does. 
We validated the proposed method via real-world experiments, where a quadruped robot is capable of traversing different terrains and resisting external disturbances.
Moreover, compared to the learning position control method, we showed that the learning torque control framework can potentially achieve a higher expected return and result in more robust locomotion behaviors. Finally, we provided discussions on the key features of learning torque control.

\section*{Acknowledgments}\label{sec:acknowledge}
We would like to thank Professor Stuart Russell and Professor Pieter Abbeel for their enlightening feedback during the course of this project, and Ayush Agrawal for his help with the experimental setup. We would also like to thank Tingnan Zhang, Xingye Da and Jonathan Li for their insightful discussions on RL-based locomotion.
This work was partially supported through a BAIR-Meta AI collaboration and Hong Kong Center for Logistics Robotics.

\newpage

\bibliographystyle{IEEEtranS}
\bibliography{references}
\balance
\end{document}